\documentclass[11pt]{article}
\usepackage[a4paper,margin=1in]{geometry}
\usepackage[T1]{fontenc}
\usepackage{lmodern,microtype}
\usepackage{amsmath,graphicx,booktabs,multirow,xcolor}
\usepackage[numbers,sort&compress]{natbib}
\usepackage{authblk}
\usepackage[font=small,labelfont=bf]{caption}
\usepackage[hidelinks]{hyperref}

\title{Where Does Free-Text 3D Grounding Fail?\\An Error Analysis of Report-Sentence Segmentation in Chest CT}

\author[1]{Abdullah Arif Sheikh}
\author[2]{Kamalaker Dadi\thanks{Corresponding author: \href{mailto:kamalaker.dadi@research.iiit.ac.in}{kamalaker.dadi@research.iiit.ac.in}}}
\author[2]{Raju S. Bapi}
\affil[1]{Department of Artificial Intelligence, Vidya Jyothi Institute of Technology, Hyderabad, India}
\affil[2]{IHub-Data, International Institute of Information Technology, Hyderabad, India}
\date{}

\begin{document}
\maketitle

\begin{abstract}
Free-text 3D grounding---segmenting, in a CT volume, the region described by one sentence of the radiology report---could link reports to images at scale, but it is unclear where current models fail.
We analyse the publicly released VoxTell model on all 381 findings of the 200-scan ReXGroundingCT validation set, where it grounds 53.5\% of findings.
Its failures are rarely about \emph{which} lesion or side is meant. No miss lies on the side opposite to the one named, and findings are almost never confused with one another.
Instead, predictions are too large, land on a neighbouring lesion on the correct side, or are missing for small findings, and diffuse findings are rarely grounded.
Removing location words halves performance, to the level of label-only prompting, and anatomical rules derived from the sentence never improve the full-sentence output.
The same pattern holds on the hidden 300-scan test set.
Further progress therefore depends on extent estimation and small-lesion sensitivity rather than on anatomical post-processing.
\end{abstract}


\section{Introduction}
\label{sec:intro}
\begin{figure}[!ht]
\centering
\includegraphics[width=0.6\textwidth]{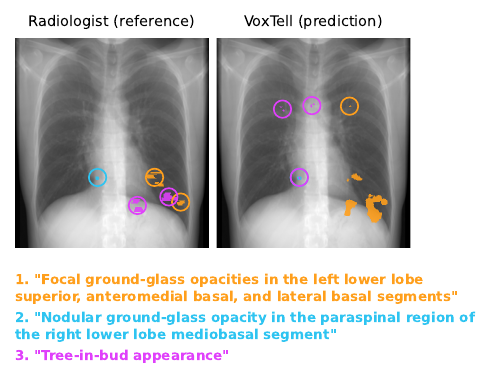}
\caption{\textbf{Free-text 3D grounding.} Each sentence of a chest CT report is a separate prompt, and the model must return a 3D mask of exactly the region that sentence describes. Colours link three sentences of one validation report to the radiologist's reference masks and to VoxTell's predictions (coronal projections, rings mark small findings). The same scan is analysed in Fig.~\ref{fig:errors}(a)--(d$'$).}
\label{fig:task}
\end{figure}

Radiology reports describe findings in free text, \emph{``two nodules, each 2\,mm, in the laterobasal segment of the left lower lobe''}, without marking where in the volume they are.
Free-text 3D grounding asks a model to segment exactly the region such a sentence refers to.
Clinically, this could let readers jump from a report sentence to the lesion it describes, support comparison of reported findings at follow-up, flag sentences without matching image evidence, and turn report archives into voxel-level labels without manual contouring~\cite{baharoon2026rexgroundingct}.
Unlike fixed-class segmentation (e.g., nnU-Net~\cite{isensee2021nnunet}) or label-prompted models that accept a short category name (SAT~\cite{zhao2025sat}, SegVol~\cite{du2024segvol}, BiomedParse~\cite{zhao2025biomedparse}), the prompt carries the type, location, number, size and descriptors of a finding, and the model must resolve all of them in 3D (Fig.~\ref{fig:task}).

ReXGroundingCT~\cite{baharoon2026rexgroundingct} made this task measurable by pairing report sentences of 3,142 non-contrast chest CTs from CT-RATE~\cite{hamamci2026ctrate} with radiologist-drawn 3D masks.
Label-prompted foundation models largely fail on it~\cite{baharoon2025mlhc}, whereas VoxTell~\cite{rokuss2026voxtell}, which embeds free text with a large language-model encoder, leads the public benchmark\footnote{\url{https://rexrank.ai/ReXGroundingCT/index.html}}.
A two-stage method that first proposes lesions and then matches them to sentences~\cite{uhm2026dagg} reports that the matching step accounts for most of its gain, suggesting that \emph{assignment} of findings to lesions is the main difficulty.
Yet no study has measured, finding by finding, how the best free-text model actually fails.

We address this gap with a controlled error analysis of the released VoxTell model on ReXGroundingCT. Our contributions are:
\begin{itemize}
\setlength\itemsep{0pt}
\item A reproducible evaluation protocol, including a mask-alignment check that this dataset requires (Sec.~\ref{sec:protocol}).
\item A finding-level error taxonomy, stratified by lesion size, finding type, focality and stated side (Sec.~\ref{sec:taxonomy}).
\item Prompt ablations and training-free post-processing showing that the model already exploits location words, so the remaining errors concern extent and small lesions rather than assignment (Sec.~\ref{sec:results}).
\end{itemize}

\section{Materials and Methods}
\label{sec:methods}

\subsection{Data and evaluation protocol}
\label{sec:protocol}
We use the MICCAI~2026 challenge release of ReXGroundingCT~\cite{baharoon2026rexgroundingct}, with 2,992 training scans, 200 validation scans with released masks (381 findings) and 300 test scans with hidden masks (582 findings). Validation and test scans are held out from the training split.
Each finding is one report sentence with a category (focal, i.e.\ discrete lesions, or non-focal, i.e.\ diffuse patterns such as emphysema) and a 3D mask that may contain several lesions (\emph{instances}). Validation masks are annotated more exhaustively than training masks, with 3.71 versus 1.95 instances per finding (1,415 instances in total).
Validation findings span five orders of magnitude in volume, from 17 voxels to over 3 million (median 3,410 voxels, interquartile range 369--22,790), and 36\% are small ($<$1,000 voxels).

\textbf{Metrics.} We follow the challenge evaluation protocol~\cite{baharoon2026rexgroundingct}.
For each finding we compute the Dice coefficient between the predicted and reference masks and count the finding as grounded (a \emph{hit}) when Dice~$\geq 0.1$, a lenient threshold that rewards locating the described region rather than matching its exact extent. Case-level Dice, used for the leaderboard ranking, is the mean over a scan's findings.
Because one finding can span several lesions, we also split masks into connected components (18-connectivity), match predicted and reference components one-to-one at Dice~$\geq 0.2$, and report instance precision (predicted lesions that match a reference lesion), recall (reference lesions that are found) and F1.
95\% confidence intervals (CIs) come from 2,000 bootstrap resamples of the 200 scans, so that findings from the same patient are resampled together; differences between settings use paired resamples.

\textbf{Mask alignment.} Masks are stored with an identity affine in the CT's native voxel order, one channel per finding, whereas the CTs carry an LPS header.
Reorienting the CT (e.g., to RAS, as VoxTell does) without applying the same transform to the mask rotates the mask by $180^\circ$ in-plane, placing, e.g., a consolidation mask on aerated lung (mean $-613$ vs.\ $-158$\,HU in native alignment).
We keep masks in native order and map predictions back through the inverse reorientation.

\subsection{Model and inference}
VoxTell~v1.1~\cite{rokuss2026voxtell} couples a 3D convolutional image encoder with a frozen Qwen3-Embedding-4B text encoder~\cite{zhang2025qwen3emb} and multi-scale vision--language fusion.
It was trained on 158 public datasets (over 62K volumes). According to its authors, the ReXGroundingCT training split was part of this training data but was sampled only rarely, because the checkpoint targets general semantic segmentation (M.~Rokuss, personal communication).
The ReXGroundingCT results in the VoxTell paper and the public leaderboard entry come from a separate variant, further fine-tuned on the ReXGroundingCT training split and evaluated with test-time augmentation, whose weights are not public~\cite{rokuss2026voxtell}.
We evaluate the publicly released weights as distributed (model checkpoint v1.1, \texttt{voxtell} package v0.1.2), without any training of our own.
We use the official sliding-window inference without test-time augmentation or post-processing, embedding and grounding each sentence independently.
Inference takes 10--60\,s per scan on a 12\,GB RTX~3080~Ti.

\subsection{Error taxonomy}
\label{sec:taxonomy}
Each validation finding is assigned the first matching type:
\emph{hit} (Dice $\geq 0.1$);
\emph{empty} (no voxel predicted);
\emph{swapped} (another sentence's prediction in the same scan hits this lesion);
\emph{partial} (prediction overlaps the lesion but Dice $<0.1$);
\emph{wrong side} (no overlap, sentence names one side, $\geq 70\%$ of the prediction lies on the other side of the body midline);
\emph{elsewhere} (all remaining misses).
We stratify by reference volume (small $<10^3$, medium $10^3$--$10^4$, large $>10^4$ voxels), by finding group derived from sentence keywords (nodule, opacity/consolidation, atelectasis/scar, pleural, airway/emphysema, other), by focality, and by the side named in the sentence (left, right, both, none).
To check this keyword parsing, we compared the side named in each sentence with the side on which its reference mask lies. They agreed for 175 of the 176 findings that name a single side.

\subsection{Prompt ablations and text-derived post-processing}
\label{sec:fixes}
To test whether the model \emph{uses} location words, we re-run inference with side, lobe, segment and region phrases removed, and with a short finding label only (e.g., ``lung nodule''), as a label-prompted model would receive.
We also test whether the anatomy named in a sentence can correct the full-sentence output without retraining. The three rules remove predicted voxels on the side opposite to the one named, remove voxels outside the named lobes (TotalSegmentator lobe masks~\cite{wasserthal2023totalseg}, keeping a component if at least a fraction $t$ of it lies inside them), and keep only as many connected components as the sentence states.
Rule thresholds are chosen on the 50 scans of the original public validation split and tested on the other 150.

\section{Results}
\label{sec:results}

\textbf{Overall.} Over all 381 validation findings, VoxTell grounds 53.5\% (95\% CI 48.6--58.4) with a mean Dice of 0.225 (95\% CI 0.203--0.250; Table~\ref{tab:strata}).
When a predicted lesion does correspond to a marked one, it is segmented well (Dice 0.517). However, only about one in eight lesions that VoxTell predicts corresponds to a lesion the radiologist marked (instance precision 0.131), and it finds about one in four of the marked lesions (recall 0.256).

\begin{table}[!ht]
\centering
\caption{VoxTell on the ReXGroundingCT validation set (200 scans, 381 findings), stratified. Hit (\%) = Dice $\geq 0.1$. Empty (\%) = no voxel predicted. Last two columns: hit rate with location words removed and with a label-only prompt. The 12 findings of the ``other'' group are included in the totals but not shown. Over the 200 validation scans, Dice per case (the metric that ranks the leaderboard) is 0.234, and instance precision, recall and F1 are 0.131, 0.256 and 0.174.}
\label{tab:strata}
\small
\setlength{\tabcolsep}{4pt}
\begin{tabular}{llrrrrrr}
\toprule
 & & & \multicolumn{3}{c}{Full sentence} & \multicolumn{2}{c}{Hit (\%)} \\
\cmidrule(lr){4-6}\cmidrule(lr){7-8}
 & Stratum & $n$ & Hit & Dice & Empty & No loc. & Label \\
\midrule
 & All & 381 & 53.5 & 0.225 & 13.1 & 27.6 & 27.6 \\
\midrule
\multirow{3}{*}{\rotatebox{90}{Size}} & Small ($<10^3$) & 137 & 43.1 & 0.192 & 24.1 & 24.1 & 30.7 \\
 & Medium & 100 & 42.0 & 0.171 & 9.0 & 19.0 & 22.0 \\
 & Large ($>10^4$) & 144 & 71.5 & 0.295 & 5.6 & 36.8 & 28.5 \\
\midrule
\multirow{5}{*}{\rotatebox{90}{Group}} & Nodule & 155 & 53.5 & 0.237 & 20.0 & 32.9 & 38.7 \\
 & Opacity/consolid. & 57 & 71.9 & 0.307 & 1.8 & 40.4 & 29.8 \\
 & Atelectasis/scar & 87 & 56.3 & 0.201 & 13.8 & 8.0 & 6.9 \\
 & Pleural & 29 & 69.0 & 0.332 & 6.9 & 55.2 & 58.6 \\
 & Airway/emphysema & 41 & 14.6 & 0.052 & 7.3 & 14.6 & 7.3 \\
\midrule
\multirow{2}{*}{\rotatebox{90}{Foc.}} & Focal & 329 & 59.0 & 0.249 & 13.7 & 29.2 & 29.2 \\
 & Non-focal & 52 & 19.2 & 0.076 & 9.6 & 17.3 & 17.3 \\
\midrule
\multirow{4}{*}{\rotatebox{90}{Side}} & Left & 76 & 68.4 & 0.323 & 3.9 & 30.3 & 35.5 \\
 & Right & 100 & 63.0 & 0.271 & 14.0 & 15.0 & 26.0 \\
 & Both & 168 & 43.5 & 0.167 & 14.9 & 34.5 & 27.4 \\
 & None & 37 & 43.2 & 0.166 & 21.6 & 24.3 & 16.2 \\
\bottomrule
\end{tabular}
\end{table}

\begin{figure}[!ht]
\centering
\includegraphics[width=0.6\textwidth]{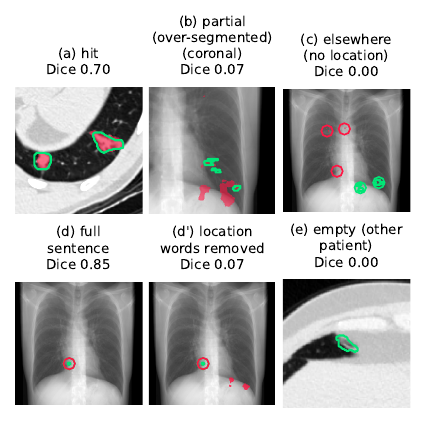}
\caption{\textbf{How VoxTell fails} (green = radiologist, red = VoxTell, rings mark small findings). Panels (a)--(d$'$) show the scan of Fig.~\ref{fig:task} and (e) another patient. (a) Hit, consolidation (Dice 0.70). (b) \emph{Partial} miss, with ground-glass opacities over-segmented $6.4\times$ (Dice 0.07). (c) \emph{Elsewhere} miss for ``tree-in-bud appearance'', which names no location. (d, d$'$) A 287-voxel nodule prompted in full (Dice 0.85) and without location words (Dice 0.07). (e) \emph{Empty} prediction for ``linear opacities in the medial segment of the right middle lobe''. Further examples of each error type, including swapped misses, are given in Appendix~\ref{app:examples}; wrong-side misses do not occur.}
\label{fig:errors}
\end{figure}

\textbf{Error types.} VoxTell misses 177 of the 381 findings. Of these, about a third each are partial (61) and elsewhere (61) misses, 50 receive no prediction at all, and only 5 are swapped. None lies on the wrong side (Fig.~\ref{fig:errors}).
In most partial misses, the prediction overlaps the lesion but is far too large, with a median predicted volume $5.3\times$ that of the reference, and 62\% exceed twice the reference. Such over-segmentation is most common for airway/emphysema and atelectasis/scar findings.
Elsewhere misses are mostly nodules (28 of 61), for which VoxTell segments a nodule on the correct side, but not the one described.
Empty predictions concentrate in small findings (24.1\% vs.\ 5.6\% for large ones).

\textbf{Strata.} Non-focal findings are grounded far less often than focal ones (19.2\% vs.\ 59.0\%), with airway and emphysema findings worst (14.6\%, Table~\ref{tab:strata}).
Sentences naming a single side are grounded more often (63--68\%) than bilateral or unlocalised ones (43\%), which are typically diffuse or multi-instance.

\begin{table}[!ht]
\centering
\caption{Prompt ablation on all 381 validation findings (top) and text-derived post-processing of full-sentence predictions on the 150 held-out scans (bottom). Wrong side = misses whose prediction lies on the side opposite to the one named. $\Delta$Dice: paired patient-level bootstrap 95\% CI against no post-processing. The lobe rule keeps a component if a fraction $\geq t$ of it lies in the named lobes.}
\label{tab:ablation}
\small
\setlength{\tabcolsep}{5pt}
\begin{tabular}{lrrrrr}
\toprule
Prompt & Hit (\%) & Dice & F1$_\text{inst}$ & Empty (\%) & Wrong side \\
\midrule
Full sentence & 53.5 & 0.225 & 0.174 & 13.1 & 0 \\
Location words removed & 27.6 & 0.104 & 0.122 & 30.2 & 33 \\
Label only & 27.6 & 0.123 & 0.122 & 31.2 & 19 \\
\midrule
Post-processing rule & Hit (\%) & Dice & \multicolumn{3}{l}{$\Delta$Dice 95\% CI} \\
\midrule
None & 55.6 & 0.230 & \multicolumn{3}{l}{--} \\
Lobe ($t=0.25$) & 55.6 & 0.231 & \multicolumn{3}{l}{[$-0.001$, $+0.001$]} \\
Lobe ($t=0.5$) & 53.8 & 0.223 & \multicolumn{3}{l}{[$-0.016$, $-0.000$]} \\
Count & 55.3 & 0.229 & \multicolumn{3}{l}{[$-0.005$, $+0.001$]} \\
\bottomrule
\end{tabular}
\end{table}

\textbf{Prompt ablations.} Removing location words halves both the hit rate (53.5\% to 27.6\%, paired 95\% CI of the change $-31.3$ to $-21.0$ points) and Dice (0.225 to 0.104, $-0.144$ to $-0.100$; Table~\ref{tab:ablation}). Label-only prompts reach the same hit rate (27.6\%, Dice 0.123), so on the same model the full sentence is worth about 26 points of hit rate over a category label (95\% CI 21.1--31.0; Dice $+0.080$ to $+0.124$).
Without location words, empty predictions more than double (13.1\% to 30.2\%) and 33 wrong-side misses appear, where the full sentence produced none.
The loss is largest where the image alone is ambiguous, such as atelectasis and scarring (56.3\% to 8.0\%), nodules (53.5\% to 32.9\%) and right-sided findings (63.0\% to 15.0\%). It is small for pleural findings, whose location is evident from the image (69.0\% to 55.2\%), and absent for airway and emphysema findings, which were already rarely grounded (14.6\% in both cases; Table~\ref{tab:strata}).

\textbf{Post-processing.} None of the three rules improves full-sentence predictions. The side rule leaves Dice unchanged on all 381 findings (0.225 to 0.225, 95\% CI of the difference 0.0000 to 0.0002), because no prediction lies on the wrong side for it to remove.
On the development scans no lobe or count setting improved Dice, and on the 150 held-out scans the lobe rule is neutral with a lenient threshold and harmful with a strict one, while count-based pruning does not raise Dice (Table~\ref{tab:ablation}). The removed components often contain true lesion voxels, because reference masks extend beyond the named lobe or contain more lesions than the sentence states.
The side rule does help once location words are removed (Dice 0.104 to 0.110, 95\% CI of the gain 0.002 to 0.011), confirming that the full sentence already supplies this information.

\textbf{Hidden test set.} On the official 300-scan test leaderboard (582 findings), the released model reaches a mean Dice per case of 0.210, a leaderboard hit rate of 48.5\% and instance precision, recall and F1 of 0.080, 0.250 and 0.122, against 0.362--0.375 Dice per case for the three top entries, which add task-specific training (leaderboard accessed November 2026).
The validation pattern holds on the test set. Pooled over the leaderboard's finding categories, non-focal findings are grounded in 20.5\% of cases (15 of 73) and focal findings in 52.5\% (267 of 509), and emphysema is again the hardest category (1 of 27 findings grounded).

\section{Discussion}
\label{sec:discussion}
Three observations reframe what limits free-text grounding.
First, the model genuinely reads location. Without location words it halves its performance and makes wrong-side errors, and with them anatomical post-processing is redundant. The matching bottleneck reported for two-stage pipelines~\cite{uhm2026dagg} therefore does not carry over.
Second, the dominant errors concern \emph{extent}. Predictions are often several times too large, and the low instance precision (0.131) shows many spurious components. For non-focal findings the model segments the whole pattern (e.g., all thickened bronchial walls), while reference masks mark representative regions, so such findings may need finding-type-aware evaluation.
Third, small findings are often missed altogether, which calls for better small-lesion sensitivity rather than better language understanding.

\textbf{Limitations.} We analyse a single model in its released form. The authors' variant, fine-tuned on ReXGroundingCT and evaluated with test-time augmentation, reports 28.2\% Dice on the earlier 50-scan public validation split~\cite{rokuss2026voxtell}, against 21.4\% for the released weights on the same 50 scans in our evaluation, and its errors may differ. Finding groups are keyword-derived, lobe masks are automatic, and CT-RATE comes from a single institution. Repeating the analysis for other grounding models and on external, multi-centre data, including Indian populations, is future work.

\section{Conclusion}
\label{sec:conclusion}
VoxTell, the leading free-text grounding model for chest CT, already uses the location given in a report sentence. It never segments the wrong side and rarely confuses one sentence with another. Its remaining errors are predictions that are too large, that land on a nearby lesion, or that miss small findings altogether. Future work should therefore improve how models estimate lesion extent and detect small lesions, rather than add anatomical rules.

\section*{Code availability}
The inference, evaluation and analysis code is available from the corresponding author on request.

\section*{Ethics}
This study used only public, de-identified data (CT-RATE and ReXGroundingCT) under their CC BY-NC-SA licences, for which ethical approval was not required.

\section*{Acknowledgements}
We thank the curators of CT-RATE and ReXGroundingCT, and Maximilian Rokuss for clarifying the training data and evaluation of VoxTell. Experiments were run on a GPU workstation at IHub-Data, IIIT Hyderabad. An AI assistant (Claude, Anthropic) helped write analysis code and draft the manuscript. The authors checked all experiments and results and take full responsibility for the content. The authors have no relevant financial or non-financial interests to disclose.

\bibliographystyle{unsrtnat}
\bibliography{refs}

\clearpage
\appendix
\section{Further examples per error type}
\label{app:examples}
\begin{figure}[h]
\centering
\includegraphics[width=\textwidth]{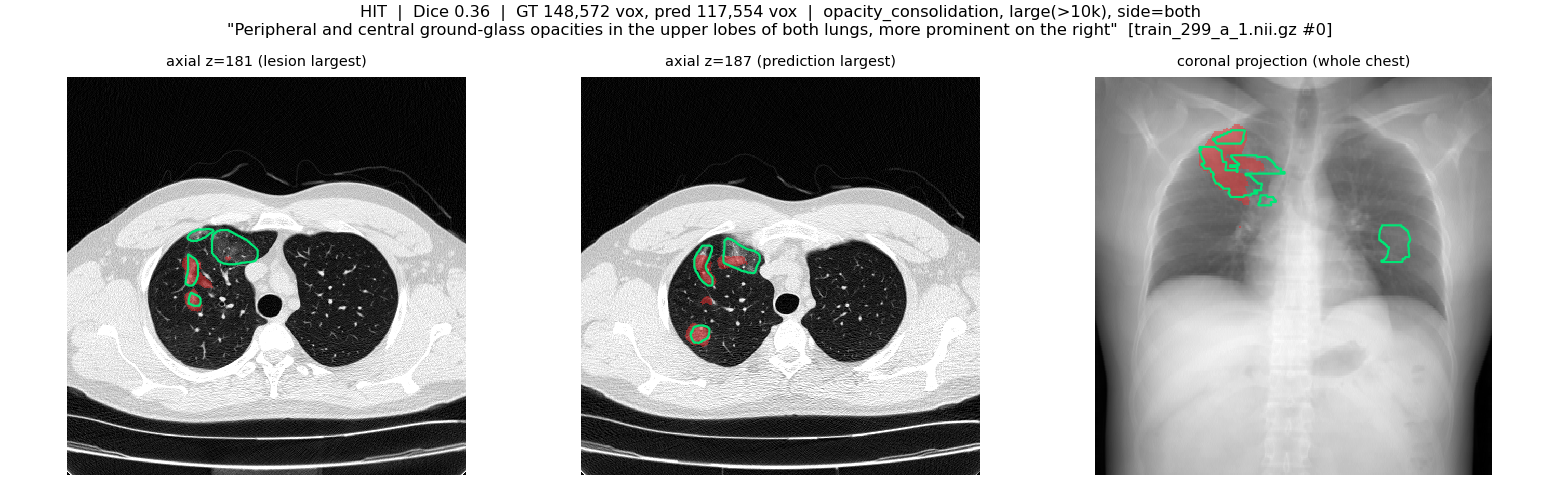}\\[4pt]
\includegraphics[width=\textwidth]{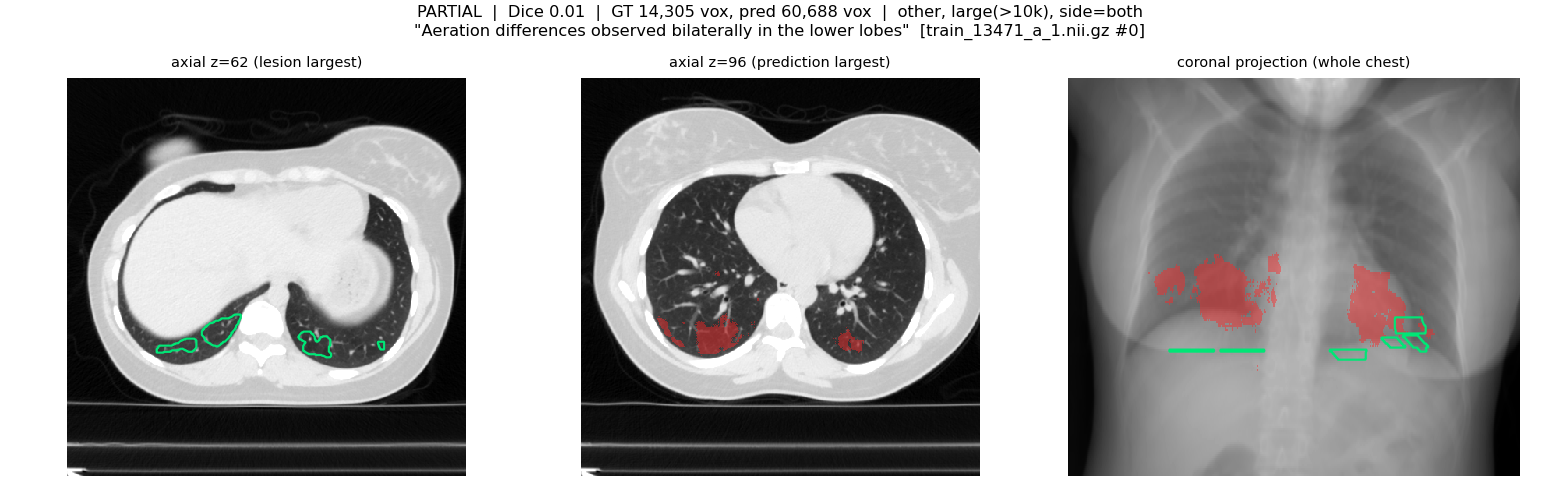}\\[4pt]
\includegraphics[width=\textwidth]{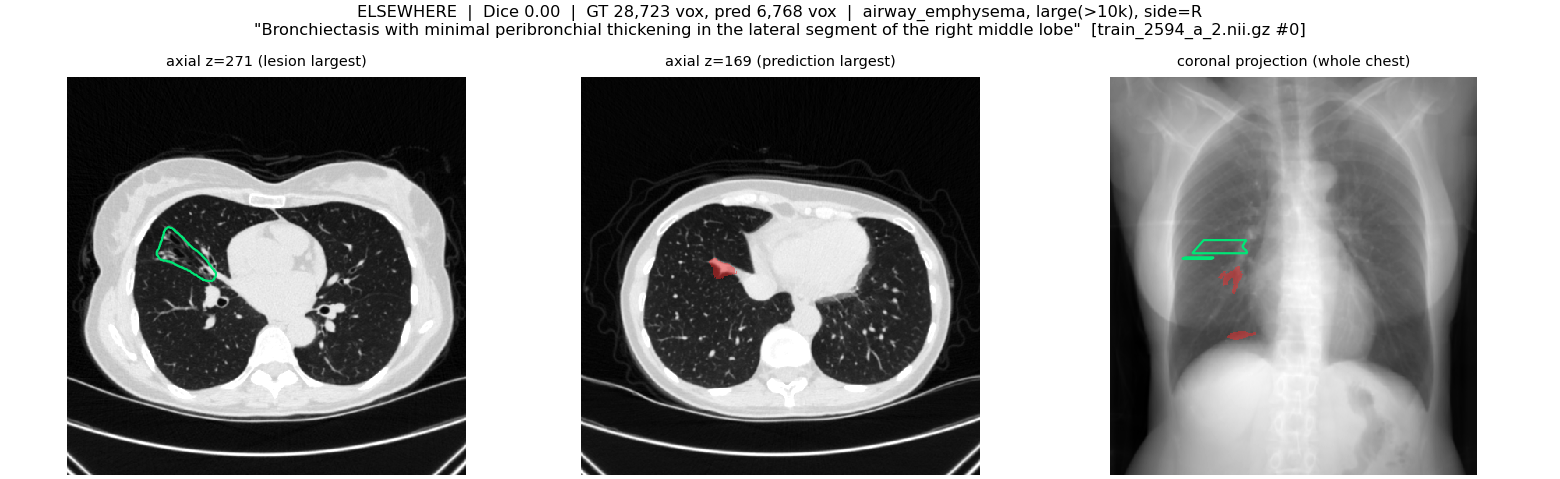}
\caption{\textbf{Hit, partial and elsewhere.} Top: a hit for bilateral upper-lobe ground-glass opacities (Dice 0.36), with the prediction confined to the right lung, where the opacities are most prominent. Middle: a \emph{partial} miss for bilateral aeration differences in the lower lobes, where the prediction overlaps the lesion but is four times its volume (Dice 0.01). Bottom: an \emph{elsewhere} miss for bronchiectasis in the right middle lobe, segmented at the right hilum instead.}
\label{fig:ex1}
\end{figure}

\begin{figure}[h]
\centering
\includegraphics[width=\textwidth]{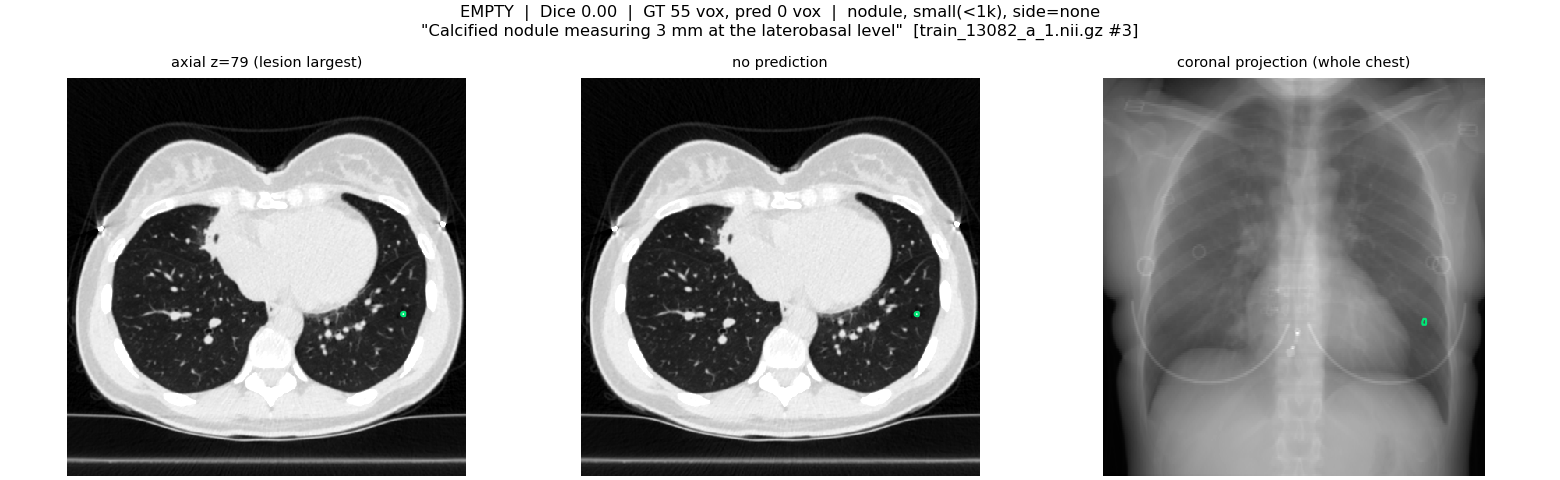}\\[4pt]
\includegraphics[width=\textwidth]{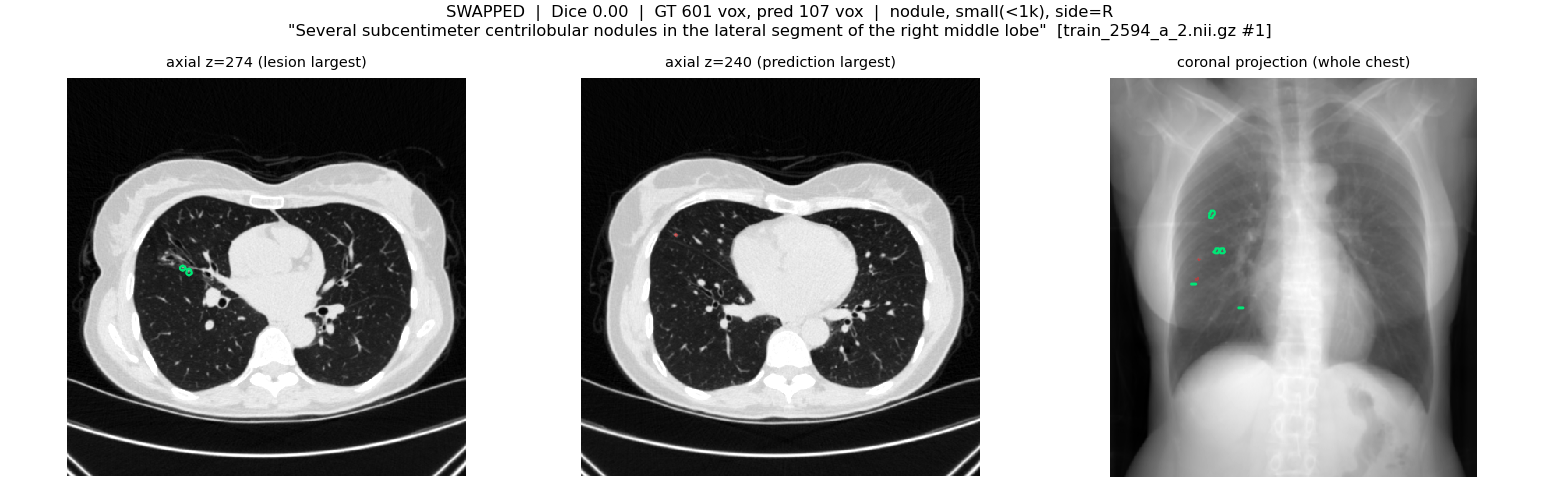}\\[4pt]
\includegraphics[width=\textwidth]{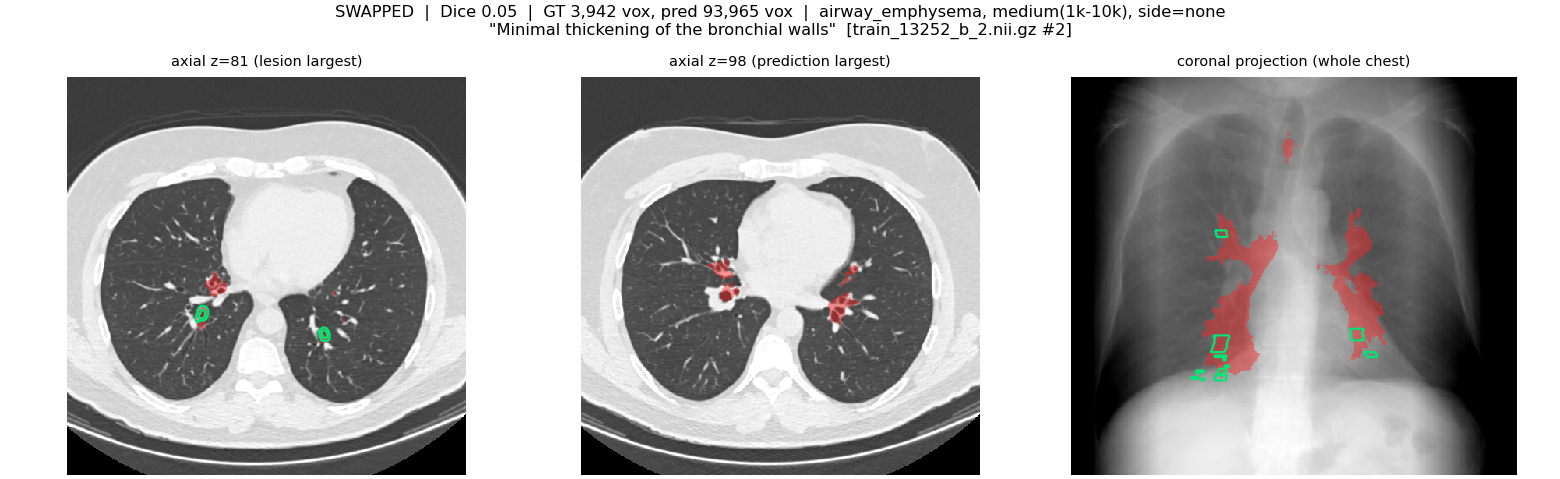}
\caption{\textbf{Empty and swapped.} Top: no prediction for a 3\,mm calcified nodule (55 voxels). Middle and bottom: \emph{swapped} misses, in which the finding's own prediction fails but the prediction for another sentence of the same scan overlaps this lesion (Dice $\geq 0.1$). The panels show the finding's own prediction, i.e.\ several right-middle-lobe nodules missed (middle) and minimal bronchial wall thickening over-segmented along both hila (bottom).}
\label{fig:ex2}
\end{figure}

Figures~\ref{fig:ex1} and~\ref{fig:ex2} show one validation finding per error type, from scans other than those in Figs.~\ref{fig:task} and~\ref{fig:errors}. Each row shows the axial slice where the reference lesion is largest, the axial slice where the prediction is largest, and a coronal average projection of the whole chest (green outline = radiologist, red = VoxTell; patient's right on the image left). The header gives the error type, Dice, reference and predicted volumes, finding group, size tier and named side, followed by the sentence.

\end{document}